\documentclass[11pt]{article}

\usepackage[preprint]{acl}

\usepackage{times}
\usepackage{latexsym}

\usepackage[T1]{fontenc}
\usepackage[utf8]{inputenc}
\usepackage{microtype}
\usepackage{inconsolata}

\usepackage{graphicx}

\usepackage{hyperref}       
\usepackage{url}            
\usepackage{booktabs}       
\usepackage{amsfonts}       
\usepackage{nicefrac}       
\usepackage{microtype}      
\usepackage{xcolor}         
\usepackage{enumitem}
\usepackage{xspace}
\usepackage{todonotes}
\usepackage{courier}
\usepackage[most]{tcolorbox}
\usepackage{multirow}
\usepackage{listings}
\usepackage[altpo,epsilon]{backnaur}

\usepackage[table]{xcolor}

\usepackage{pifont}
\newcommand{\cmark}{\textcolor{green}{\ding{51}}} 
\newcommand{\xmark}{\textcolor{red}{\ding{55}}}

\newcommand{\DatasetFramework}{SIMPACT\xspace}

\usepackage{amssymb}
\usepackage{pifont}

\title{\texttt{BluePrint}: \\ A Social Media User Dataset for LLM Persona Evaluation and Training}

\author{
 \textbf{Aurélien Bück-Kaeffer\textsuperscript{1, 2}},
 \textbf{Je Qin Chooi\textsuperscript{3}},
 \textbf{Dan Zhao\textsuperscript{4}},
 \textbf{Maximilian Puelma Touzel\textsuperscript{1,5}},
\\
 \textbf{Kellin Pelrine\textsuperscript{1,2}},
 \textbf{Jean-François Godbout\textsuperscript{2, 5}},
 \textbf{Reihaneh Rabbany\textsuperscript{1, 2}},
 \textbf{Zachary Yang \textsuperscript{1,2}},
\\
\\
 \textsuperscript{1}McGill University,
 \textsuperscript{2}Mila - Quebec Artificial Intelligence Institute,
 \\
 \textsuperscript{3}Harvard College,
 \textsuperscript{4}NYU,
 \textsuperscript{5}Université de Montréal
\\
}

\begin{document}
\maketitle


\begin{abstract}

Large language models (LLMs) offer promising capabilities for simulating social media dynamics at scale, enabling studies that would be ethically or logistically challenging with human subjects. However, the field lacks standardized data resources for fine-tuning and evaluating LLMs as realistic social media agents. We address this gap by introducing \textbf{\DatasetFramework}, the \textbf{SIM}ulation-oriented \textbf{P}ersona and \textbf{A}ction \textbf{C}apture \textbf{T}oolkit, a privacy respecting framework for constructing behaviorally-grounded social media datasets suitable for training agent models. We formulate next-action prediction as a task for training and evaluating LLM-based agents and introduce metrics at both the cluster and population levels to assess behavioral fidelity and stylistic realism. As a concrete implementation, we release \texttt{BluePrint}, a large-scale dataset built from public Bluesky data focused on political discourse. \texttt{BluePrint} clusters anonymized users into personas of aggregated behaviours, capturing authentic engagement patterns while safeguarding privacy through pseudonymization and removal of personally identifiable information. The dataset includes a sizable action set of 12 social media interaction types (likes, replies, reposts, \textit{etc.}), each instance tied to the posting activity preceding it. This supports the development of agents that use context-dependence, not only in the language, but also in the interaction behaviours of social media to model social media users.  
By standardizing data and evaluation protocols, \DatasetFramework provides a foundation for advancing rigorous, ethically responsible social media simulations. \texttt{BluePrint} serves as both an evaluation benchmark for political discourse modeling and a template for building domain-specific datasets to study challenges such as misinformation and polarization.
\end{abstract}

\section{Introduction}

Social media platforms have become critical spaces for public discourse, collective decision-making, and social interaction. Recent advances in large language models (LLMs) present new opportunities for simulating these environments through AI agents \cite{touzel2024simulation, yang2025oasisopenagentsocial}. By modeling social media users as LLM-based agents, researchers can build interactive, scalable simulations to study phenomena such as information diffusion, community formation, and platform interventions. These simulations offer controlled, reproducible testbeds for investigating complex dynamics like misinformation, polarization, and coordinated manipulation—challenges that are often impractical or ethically risky to study with human participants.

Despite their promise, LLM-based social media simulations face key limitations. Current approaches often rely on simplified scenarios that fail to capture the complexity of real user behavior. The absence of standardized benchmarks makes evaluation inconsistent and progress difficult to measure \cite{larooij2025largelanguagemodelssolve}. Moreover, privacy concerns \cite{king2021new, isaak2018user} restrict access to authentic user data, limiting the realism. These gaps hinder the development of robust, behaviorally grounded social media agents.

\begin{figure*}[t!]
\centering
\includegraphics[width=0.99\linewidth]{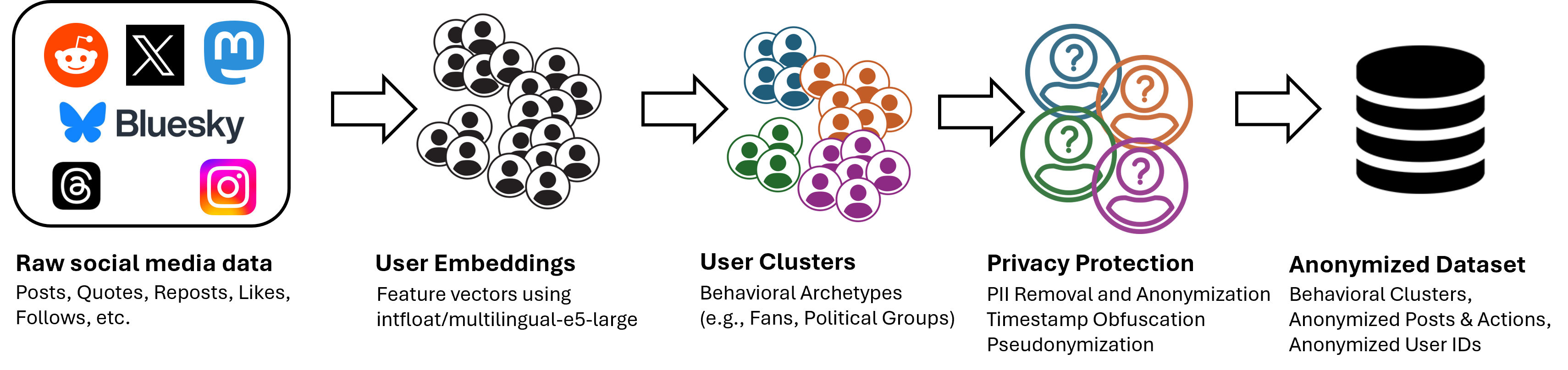}
\caption{\label{fig:dataset_creation_process_diagram} \textbf{\DatasetFramework (Simulation-oriented Persona and Action Capture Toolkit)} for constructing privacy-preserving social media user datasets. We collect raw social media data, generate user embeddings, and cluster users into behavioral archetypes (e.g., fans, political groups) using constrained K-means. Privacy  is preserved through PII removal, timestamp obfuscation, and per-thread pseudonymization of user IDs. The resulting dataset contains anonymized text, actions, and behavioral clusters suitable for social media agent simulation.}
\end{figure*}

We address these challenges by introducing \DatasetFramework, a generalizable, privacy-preserving framework for constructing datasets to train and evaluate LLM-based social media agents (Figure~\ref{fig:dataset_creation_process_diagram}). Central to our framework is the task of \textit{next-action prediction}, which frames user behavior as sequences of actions—such as posting, liking, replying, and following. To balance realism and privacy, we cluster users into \textit{behavioral personas} that capture diverse interaction behaviors while abstracting individual identities. This enables the creation of ethically responsible datasets that preserve social dynamics without exposing sensitive user data.

To demonstrate the utility of our framework, we present \texttt{BluePrint}, a large-scale, publicly available dataset constructed from Bluesky data. Focusing on political discourse during the 2025 Canadian federal election, \texttt{BluePrint} contains millions of user actions organized into multi-turn conversational threads. These threads include both textual content and interaction behaviors, supporting models that learn contextual engagement patterns. By clustering users at multiple granularity levels, \texttt{BluePrint} enables the simulation of diverse social media behaviors across different community archetypes. All data is fully anonymized to protect user privacy while maintaining behavioral richness.

We validate our framework by benchmarking several LLMs (GPT-4.1 mini, GPT-o3 mini, and Qwen 2.5) on the next-action prediction task using \texttt{BluePrint}. Our findings reveal that while current models (esp. fine-tuned) can generate plausible text, they struggle to reproduce the nuanced behavioral patterns of real user communities. These results underscore the need for standardized datasets and evaluation protocols in this emerging research area.

To summarize, our \textbf{principal contributions} are:
\begin{itemize}[nosep]
    \item \textbf{\DatasetFramework:} A privacy-preserving framework for constructing behaviorally rich, thread-based datasets to train and evaluate LLM-based social media agents.
    \item \textbf{\texttt{BluePrint}:} A publicly available, persona-based dataset derived from Bluesky, designed to benchmark social media agents in political discourse simulations.
    \item \textbf{Benchmarks and Analysis:} Empirical evaluation of state-of-the-art LLMs, highlighting their strengths and limitations in simulating user behavior across behavioral clusters.
\end{itemize}

\section{Related Works}

Recent research highlights the potential of LLMs to simulate social media dynamics across various domains. 
While several existing datasets support persona modeling or user behavior simulation, they typically focus on dialogue generation, lack interaction diversity, or do not implement privacy safeguards. Table~\ref{tab:related_datasets} summarizes key differences between these resources and \texttt{BluePrint}, which uniquely offers large-scale, privacy-preserving interaction data spanning both text- and user-directed actions.

\begin{table*}[htbp!]
\centering
\scriptsize
\begin{tabular}{p{2.5cm} p{3.5cm} p{2.5cm} c c c c c}
\hline
Dataset &
  Description &
  Platform &
  Posts &
  Action$_{text}$ &
  Action$_{all}$ &
  Privacy &
  Public \\
\hline
Persona-Chat \cite{zhang-etal-2018-personalizing} &
  10k multi-turn dialogues w/ paired persona profiles & 
  --  &
  \cmark &
  \xmark &
  \xmark &
  $\mathord{?}$ &
  \cmark \\
PersonalDialog \cite{zheng2020personalizeddialoguegenerationdiversified} &
  20.8M dialogues tagged w/ speaker metadata &
  Weibo &
  \cmark &
  \xmark &
  \xmark &
  \xmark &
  \cmark \\
Synthetic Persona-Chat \cite{jandaghi-etal-2024-faithful} &
  AI-generated multi-turn dialogues &
  -- &
  \cmark &
  \xmark &
  \xmark &
  \cmark &
  \cmark \\
COVID-19 Vaccine Engagement \cite{qiu2025llmssimulatesocialmedia} &
  3.9K tweets, retweets, quotes on COVID-19 vaccination &
  Twitter &
  \cmark &
  \cmark &
  \xmark &
  \xmark &
  $\mathord{?}$ \\
Chirper.ai Social Network \cite{zhu2025characterizingllmdrivensocialnetwork} &
  7.7M posts across 65k AI-driven user accounts  &
  Chirper.ai &
  \cmark &
  \cmark &
  \xmark &
  \cmark &
  \xmark
   \\
FineRob \cite{li2024finegrainedbehaviorsimulationroleplaying} &
  78.6k behavior simulation records across 1,866 users &
  Twitter, Reddit, Zhihu &
  \cmark &
  \cmark &
  \cmark &
  \xmark &
  \cmark \\ \\ \hline \\ 
\textbf{BluePrint (ours)} &
  \textbf{6.8M message chains across 236k unique users} &
  \textbf{BlueSky} & 
  \cmark &
  \cmark &
  \cmark &
  \cmark &
  \cmark \\ \\
\hline
\end{tabular}
\caption{\label{tab:related_datasets} \textbf{Comparison of related datasets for persona and behavior modeling.} \texttt{BluePrint} uniquely combines large-scale, anonymized and aggregated social media interaction data with both textual and non-textual actions, supporting the development and evaluation of realistic social media agents.}
\end{table*}

\paragraph{Social Media Simulation Environments}
Recent work has explored the use of LLM agents to simulate social media dynamics at scale. For example, \cite{touzel2024simulation} paired LLM agents with Mastodon to model political manipulation, while \cite{piao2025agentsocietylargescalesimulationllmdriven} introduced AgentSociety, featuring 10,000 agents to study polarization and policy outcomes. Other efforts have focused on real-world data analysis, such as mapping global election discourse on Facebook \cite{pecile2024mappingglobalelectionlandscape} or simulating opinion leader emergence on social networks \cite{jin2024buildinfluentialbotsocial}. More general simulation platforms like OASIS \cite{yang2025oasisopenagentsocial} and RecAgent \cite{wang2024userbehaviorsimulationlarge} scale to hundreds of thousands of LLM-driven users with dynamic network structures. While these studies demonstrate the potential of LLM-based simulations, they typically rely on synthetic or scripted personas and lack consistent evaluation protocols grounded in real-world behavior. In contrast, \DatasetFramework focuses on constructing and evaluating behaviorally grounded agents using privacy-preserving clusters derived from real social media users, as exemplified by \texttt{BluePrint}.

\paragraph{Fine-Tuning and Steering LLM Personas}
Prior work has explored steering LLM outputs toward specific personas through fine-tuning on ideological or community-aligned data \cite{li2024steerabilitylargelanguagemodels, jiang2022communitylmprobingpartisanworldviews}. While these methods demonstrate that LLMs can be adapted to reflect group identities, they often rely on static datasets or single-turn prompts. In contrast, \DatasetFramework provides a multi-turn, behaviorally-grounded resource for modeling both linguistic style and context-dependent interaction patterns.

\paragraph{Character Imitation and Role-Playing}
Several studies have explored how LLMs can emulate fictional or scripted characters through role-playing. For example, ChatHaruhi \cite{li2023chatharuhirevivinganimecharacter} conditioned LLMs on character memories from anime scripts, while LIFECHOICE \cite{xu2024characterdestinyroleplayinglanguage} evaluated persona-consistent decision-making for literary figures. Neeko \cite{yu2024neekoleveragingdynamiclora} further demonstrated dynamic persona switching using LoRA-based adaptation. While these works show that LLMs can mimic predefined fictional personas, they primarily focus on narrative settings. In contrast, \DatasetFramework targets real-world social media personas derived from clustered user behavior, bridging the gap between fictional role-play and authentic social dynamics to support realistic, behaviorally grounded agent simulations.

\paragraph{Evaluation of Simulated Agents and Outputs}
Evaluating the fidelity of LLM-based agents remains a major challenge. OpinionQA \cite{santurkar2023opinionslanguagemodelsreflect} evaluates alignment between LLM-generated opinions and demographic groups, revealing persistent mismatches even with persona prompting. FineRob \cite{li2024finegrainedbehaviorsimulationroleplaying} decomposes social media behavior into QA-style tasks, providing fine-grained behavior assessments. Dialogue datasets like Persona-Chat \cite{zhang-etal-2018-personalizing} and PersonalDialog \cite{zheng2020personalizeddialoguegenerationdiversified} offer resources for persona-grounded generation, but focus on single-turn or small-scale interactions. \DatasetFramework extends these efforts by introducing population-level evaluation metrics based on behavioral clusters, assessing not just per-utterance believability but also how well agents replicate aggregate community behaviors. Unlike crowd-sourced or synthetic persona datasets, \DatasetFramework applies privacy-preserving safeguards, enabling realistic and ethical evaluation of agent behavior at scale.

\section{ \DatasetFramework } \label{sec:dataset_framework}

To advance the responsible development of LLM-based social media simulations, we propose \DatasetFramework, a generalizable framework for constructing privacy-preserving, behaviorally grounded datasets. This framework provides a systematic process for transforming raw social media data into structured resources that support agent-based modeling while minimizing the risks of user re-identification and harm. Recognizing that social media dynamics are shaped not only by user-generated content but also by interaction behaviors (\textit{e.g.} liking, following, quoting, \textit{etc.}), \DatasetFramework is designed to capture the full spectrum of user actions. 

Figure~\ref{fig:dataset_creation_process_diagram} illustrates the key components of \DatasetFramework, including user clustering and privacy protection through PII removal and anonymization, timestamp obfuscation, and user pseudonymization. Together, these processes produce datasets that provide a foundation for building and evaluating agent-based social media simulations.

\subsection{User Clustering}

To balance data utility, scalability, and privacy, we model behavioral archetypes rather than individual users. Clustering users into broader behavioral groups serves both practical and ethical goals: it increases data availability for training, reduces the risk of overfitting to individual accounts, and mitigates the ethical risks of simulating identifiable individuals without consent. This abstraction enables simulation frameworks to represent diverse interaction styles, topic engagement, and platform usage patterns using a manageable number of agent profiles, supporting large-scale simulations while preserving meaningful behavioral variation.

Inspired by recent work in political and social media user modeling \cite{yang2024regionaltemporalpatternspartisan}, we compute user embeddings by averaging sentence-level representations of all posts, quotes, and reposts authored by each user. We use the \texttt{intfloat/multilingual-e5-large} model to generate these embeddings, capturing both topical and stylistic signals. Users are then grouped using constrained K-means clustering \citep{Levy-Kramer_k-means-constrained_2018}, enforcing a minimum cluster size of 10 to avoid trivial clusters.

Qualitative inspection of the resulting clusters reveals coherent behavioral groupings, such as scientific communities, sports fandoms, and political affiliations. While not all clusters are easily interpretable, the overall structure preserves meaningful diversity across the dataset. This enables the development of agent models that generalize across user groups while minimizing privacy risks. A summary of labeled clusters is provided in Appendix~\ref{tab:appendix_tf_idf_keywords}.

\subsection{PII Removal and Anonymization} \label{sec:pii_removal}

We next take steps to remove personally identifiable information (PII) that is not essential to research goals. We use Presidio \citep{microsoft_presidio}, an open-source privacy tool, to automatically detect and replace sensitive entities such as email addresses, phone numbers, credit card details, IP addresses, cryptocurrency addresses, and URLs. For example, a message like ``We welcome feedback at \texttt{janedoe@gmail.com}'' would be transformed to ``We welcome feedback at \texttt{<EMAIL\_ADDRESS>}''.

We apply the same principle to user mentions. On Bluesky, usernames typically appear in the format \texttt{@username.bsky.social}. We anonymize these by replacing them with a generic placeholder such as \texttt{@<USERNAME>}. This also includes custom domains used as usernames, such as government or organizational accounts (e.g., \texttt{@govevers.wisconsin.gov}), which are treated in the same way. This ensures that while the dataset preserves interaction patterns and conversational structure, it does not expose identifiable user information.

\subsection{Timestamp Obfuscation} \label{sec:relative_timestamp}

To protect user privacy while preserving interaction dynamics, we obfuscate all exact timestamps. Precise Unix epoch times can potentially be cross-referenced with external data sources, posing re-identification risks. To mitigate this, we sort all messages by their original timestamps and replace them with relative sequence ranks, reassigning each message a position from 1 to $N$, where $N$ is the total number of messages within each cluster in the dataset. This preserves the overall chronological order while removing access to the original posting times. When multiple messages share the same timestamp, we resolve ties arbitrarily. This allows researchers to study realistic interaction patterns without exposing sensitive temporal metadata.

\subsection{Pseudonymization} \label{sec:pseudonymization}

To further reduce the risk of user re-identification, we apply a pseudonymization procedure consistent with GDPR principles \cite{limniotis2021cryptography}, ensuring that account identifiers in our dataset cannot be linked to real individuals without access to secret cryptographic information.

All user identifiers are replaced with cryptographically secure hashes. We generate a 32-byte secret key and use it to re-compute anonymized Distributed Identifiers (DIDs) for each conversation thread. The hashing process combines the original DID, the thread content (excluding identifiers), and the secret key, producing a consistent pseudonym for the same user \textit{within} a thread, while assigning a different pseudonym \textit{across} threads.

This design prevents cross-thread linkage, reducing the risk that re-identification in one part of the dataset could expose a user’s full activity. Because the secret key is never shared, external parties cannot reverse the hashing process to recover original account identifiers. Although Bluesky data is public, we apply this additional layer of protection to respect user privacy and to enable data removal requests. Our hashing scheme allows us to reliably identify and delete a user’s content upon request without ever exposing their original account information.

\subsection{Action Set}

We define an \textit{action} as any type of interactions that are common across most social media platforms and form the foundation of user engagement dynamics online. Different action types have different attributes. Some depend on an existing post (\textit{e.g.} \texttt{reply}), while others on an another user (\textit{e.g.} \texttt{follow}). Table~\ref{tab:actions_descriptions} presents the full set of actions represented in our dataset, categorized into two broad types: text-directed actions, which target specific posts or content, and user-directed actions, which target the authors of that content.

\begin{table*}[htbp!]
\centering
\small
\begin{tabular}{llll}
\hline
\textbf{Action} &  \textbf{Type} & \textbf{Description} \\
\hline
\texttt{post}      &  Text-directed & Create a new post  \\
\texttt{reply}      &  Text-directed  & Reply to an existing post  \\
\texttt{quote}       &  Text-directed & Quote an existing post  \\
\texttt{post\_update}  &  Text-directed & Edit or update a post  \\
\hline
\texttt{post\_delete}   &  Text-directed & Delete a post  \\
\texttt{repost}         &  Text-directed &  Repost an existing post \\
\texttt{unrepost}       &  Text-directed & Remove a previously made repost& \\
\texttt{like}            &  Text-directed & Like a post \\
\texttt{unlike}         &  Text-directed &  Remove a like from a post \\
\midrule
\texttt{follow}        &  User-directed & Follow the user who authored the post\\
\texttt{unfollow}      &  User-directed &  Unfollow the user who authored the post \\
\texttt{block}        &  User-directed & Block the user who authored the post  \\
\texttt{unblock}      &  User-directed &  Unblock the user who authored the post \\
\hline
\end{tabular}
\caption{\label{tab:actions_descriptions} \textbf{User actions captured using \DatasetFramework.} The dataset includes both text-directed actions (e.g., posting, replying, liking) and user-directed actions (e.g., following, blocking), enabling comprehensive modeling of content engagement and social network interactions for agent-based simulation.}
\end{table*}

While text-directed actions such as \texttt{like}, \texttt{reply}, and \texttt{quote} have clear references to specific posts in the dataset, user-directed actions present additional modeling challenges. For actions like \texttt{follow}, \texttt{unfollow}, \texttt{block}, and \texttt{unblock}, we have no direct visibility into the full user history or external factors that may have influenced these decisions. As such, we make a practical and consistent assumption: these actions are treated as responses to the most recent post authored by the target user at the time the action was taken. We acknowledge that this is a simplification and may not reflect the true motivation behind every user-directed action. However, this treatment ensures that all actions in the dataset can be consistently linked to visible user behavior.

\subsection{Threads}

We organize the dataset into \textit{threads}, each representing an action sequence of social media interactions. Informally, a thread begins with an initial post, followed by zero or more additional posts, and concludes with the user action (e.g., \texttt{like}, \texttt{repost}, \texttt{reply}). Conceptually, they approximate conversation threads on social platforms. In practice they are stored in JSON format. The modeling objective is to predict the final element of the thread, the next plausible action, as a proxy for simulating user behavior. Cluster membership is determined by the producer of this final element. Specifically, if the last element was authored by a user from cluster $X$, the entire thread is labeled as belonging to cluster $X$.
Formally, thread structure can be defined using Backus–Naur Form (BNF) as follows:
\begin{bnf*}
\bnfprod{thread}{\bnfpn{post}\bnfpn{posts}\bnfpn{action}}\\
\bnfprod{posts}{[\bnfpn{post}]\bnfpn{posts}\bnfor\bnfes}
\end{bnf*}

Here, $\bnfes$ represents an empty sequence, allowing threads to include any number of intermediate posts. This unambiguous formalization is meant to facilitate usage of our dataset by others.
\section{BluePrint} \label{sec:blueprint}

To demonstrate the practical application of \DatasetFramework, we introduce \texttt{BluePrint}, a curated dataset capturing political discourse surrounding the 2025 Canadian Federal Election on Bluesky. \texttt{BluePrint} applies our framework end-to-end, covering data collection, cleaning, clustering, and anonymization.

\paragraph{Data Collection}
We collected public Bluesky posts, likes, and follows from March 2025 using the official Jetstream client\footnote{\url{https://github.com/bluesky-social/jetstream}}. To focus on election-related discourse, we filtered for posts containing at least one term from a curated list of political keywords (97), including candidate handles, party identifiers (43), and general election terms (11) (Section~\ref{sec:bluesky_2025_cad_federal_election_keywords}).

\paragraph{Data Curation}
We retained only English-language posts, based on Bluesky’s language metadata, and removed users with one or less English post to limit the presence of outliers and ensure linguistic consistency across the dataset.

\paragraph{Multi-Scale User Clustering}
BluePrint provides clustering at multiple granularities ($K = 2$, 25, 100, and 1000) to support different simulation scenarios. Specifically, using silhouette score \cite{9260048}, we get a natural cluster size of 2. Larger $K$ values capture increasingly fine-grained behavioral archetypes, consistent with population sizes (max 1000) used in prior simulation work \cite{touzel2024simulation, mou2024individualsocietysurveysocial}.

\subsection{Dataset Statistics}

BluePrint comprises \textbf{6.8 million actions} from \textbf{236,331 distinct users}. Table~\ref{tab:actions_stats} details the 25 behavioral archetypes used in our experiments. Analyses for all 4 granularity levels appear in Appendix \ref{sec:cluster_stats}, with comprehensive statistics available in the dataset repository. These metrics highlight the diversity of user behaviors and cluster distributions captured in our dataset.

\begin{table*}[htbp!]
\centering
\scriptsize
\begin{tabular}{llllllll}
\hline
\textbf{Action} & Cluster 0 & Cluster 1 & ... & Cluster 23 & Cluster 24 & \textbf{Average} & \textbf{Total}\\
\hline
\texttt{post} & 34101 & 15686 & ... & 72011 & 79036 & 61960.16 & 1549004 \\
\texttt{reply} & 1794 & 220 & ... & 1609 & 2710 & 1876.52 & 46913 \\
\texttt{quote} & 224 & 123 & ... & 1001 & 3480 & 1696.48 & 42412\\
\texttt{post\_update} & 50 & 2 & ... & 145 & 57 & 40.00 & 1000\\
\texttt{post\_delete} & 0 & 0 & ... & 1 & 3 & 0.92 & 23\\
\texttt{repost} & 13 & 0 & ... & 44 & 74 & 51.12 & 1278 \\
\texttt{unrepost} & 0 & 0  & ... & 0 & 0 & 0.16 & 4 \\
\texttt{like} & 10155 & 2579 & ... & 28130 & 108601 & 63550.28 & 1588757 \\
\texttt{unlike} & 79 & 20 & ... & 245 & 780 & 437.00 & 10925 \\
\texttt{follow} & 51126 & 4223 & ... & 185512 & 166658 & 140713.96 & 3517849\\
\texttt{unfollow} & 1712 & 226 & ... & 6772 & 2688 & 2762.76 & 69069\\
\texttt{block} & 3 & 6 & ... & 8 & 42 & 32.20 & 805 \\
\texttt{unblock} & 0 & 0 & ... & 0 & 1 & 0.08 & 2\\
\hline
Total Actions & 99257 & 23085 & ... & 295478 & 364130 & 273121.64 & 6828041 \\
\hline
No. of Users & 6700 & 3062 & ... & 8901 & 11517 & 9453.24 & 236331 \\
\hline
\end{tabular}
\caption{\label{tab:actions_stats} Dataset statistics for the  25 archetypes version}
\end{table*}

To aid interpretability, we provide two additional resources: \textbf{TF-IDF Keyword Analysis} (Appendix~\ref{tab:appendix_tf_idf_keywords}) highlighting salient terms for each cluster, and \textbf{Medoid Approximation} presenting the most representative posts characterizing each cluster. These resources are included alongside our dataset\footnote{\url{https://huggingface.co/datasets/ComplexDataLab/BluePrint}}, supporting transparent exploration of cluster semantics.
\section{Methodology} \label{sec:methodolgoy}

Our objective is to facilitate social media agent development—LLM-based models that replicate both language and interaction patterns of user clusters. These agents provide a foundation for future agent-based modeling and behaviorally grounded social media simulations.

\subsection{Model setup}

We evaluate the ability of LLMs to imitate social media users using two state-of-the-art proprietary models (\texttt{GPT-4.1-mini} and \texttt{o3-mini}) alongside open-weight models based on \texttt{Qwen-2.5-7B-Instruct}. We include the base \texttt{Qwen-2.5-7B-Instruct} as a control, and train two LoRA adapters \cite{hu2021loralowrankadaptationlarge} on the \texttt{BluePrint} dataset: one using focal loss and one using standard cross-entropy loss. Training is detailed in Appendix \ref{sec:qwen_fine_tuning}.

\subsection{Evaluation metrics}

Evaluating whether an LLM convincingly imitates human social media behavior is inherently challenging. Computational metrics offer approximations but cannot fully capture the subtlety of human-like behavior, often producing uncanny imitations. To address this, we adopt a multi-metric evaluation protocol:

\begin{itemize}
\item \textbf{Maximum Cosine Similarity}: Measures the highest similarity between model-generated and ground-truth post embeddings.
\item \textbf{Average Embedding Cosine Similarity}: Compares the average embedding of all model outputs with that of ground-truth posts.
\item \textbf{Jaccard Similarity (Top-100 TF-IDF Terms)}: Measures lexical overlap between top keywords in model-generated and real posts.
\item \textbf{JS (Jensen-Shannon) Divergence}: Measures distributional similarity between model-generated and real post embeddings; lower values indicate higher similarity.
\item \textbf{F1 Score}: Measures accuracy of predicting user actions (e.g., like, follow, repost, ignore) relative to observed user behaviors.
\end{itemize}

All metrics are computed both at the cluster level (to assess subgroup fidelity) and averaged across clusters (to assess overall population-level performance). To complement these computational measures, we conduct a human evaluation in which annotators are shown pairs of real and model-generated posts and asked to identify the human-written one. An accuracy near 50\% indicates that the model is indistinguishable from real users, while higher accuracy reveals detectable gaps in behavioral realism.
\section{Experiments} \label{sec:experimental_results}

We ran the \DatasetFramework pipeline and evaluated results using our metrics, benchmarking publicly available LLMs while demonstrating our approach's effectiveness. For our experiment, we arbitrarily selected 5 clusters (0, 1, 6, 18 and 21) from the $k=25$ configuration. Results for all 5 clusters appear in Table \ref{tab:metrics_for_all_cluster_table}.

\begin{table*}[htbp!]
\centering
\tiny
\begin{tabular}{llccccc}
\hline
\textbf{Cluster} & \textbf{Metric (↑/↓)} & \textbf{GPT-4.1-mini} & \textbf{o3-mini} & \textbf{Qwen2.5-7B-Instruct} & \textbf{Qwen2.5-7B-Inst.\textsubscript{focal}} & \textbf{Qwen2.5-7B-Inst.\textsubscript{CE}} \\
\hline
\rowcolor[HTML]{F3F3F3}
\textbf{1 (Best)} & Jaccard Similarity (↑)            & 0.0101 & 0.0000 & 0.0050 & \textbf{0.1364} & 0.1299 \\
\rowcolor[HTML]{F3F3F3}
                  & Avg. Cosine Similarity (↑)        & 0.9390 & 0.9321 & 0.9341 & \textbf{0.9909} & 0.9947 \\
\rowcolor[HTML]{F3F3F3}
                  & Max. Cosine Similarity (↑)        & 0.8685 & 0.8660 & 0.8581 & \textbf{1.0000} & \textbf{1.0000} \\
\rowcolor[HTML]{F3F3F3}
                  & JS Divergence (↓)                & 0.3232 & 0.4739 & 0.5482 & 0.0680 & \textbf{0.0435} \\
\rowcolor[HTML]{F3F3F3}
                  & F1 Score (↑)                      & 0.5276 & 0.5354 & 0.5276 & 0.5354 & \textbf{0.6220} \\
\hline
\textbf{6 (Worst)} & Jaccard Similarity (↑)            & 0.0050 & 0.0152 & 0.0050 & 0.0989 & \textbf{0.1429} \\
                  & Avg. Cosine Similarity (↑)        & 0.8944 & 0.8896 & 0.8907 & \textbf{0.9293} & 0.9152 \\
                  & Max. Cosine Similarity (↑)       & \textbf{1.0000} & \textbf{1.0000} & \textbf{1.0000} & \textbf{1.0000} & \textbf{1.0000} \\
                  & JS Divergence (↓)                & 0.5340 & 0.5818 & 0.5977 & \textbf{0.4770} & 0.4973 \\
                  & F1 Score (↑)                      & \textbf{0.2077} & \textbf{0.2077} & 0.1913 & 0.2022 & 0.1257 \\
\hline
\end{tabular}
\caption{\label{tab:metrics_cluster_1_vs_cluster_6} \textbf{Cluster-level performance metrics} for multiple models imitating social media users, using \texttt{BluePrint} as ground truth. Results are shown for Cluster 1 (best-performing) and Cluster 6 (worst-performing) from the 25-cluster partition. Best scores per metric are bolded.}
\end{table*}

In Table~\ref{tab:metrics_cluster_1_vs_cluster_6}, we report multiple behavioral and embedding-based metrics. While Cluster 6 achieves perfect Max. Cosine Similarity across all models, this does not translate to higher lexical or behavioral alignment, as reflected in its low Jaccard Similarity and F1 scores. This suggests that high embedding similarity alone (e.g., a single maximally similar sample) may overestimate model alignment without capturing broader behavioral patterns. In contrast, Cluster 1 shows stronger overall alignment across metrics, including higher Jaccard Similarity, lower JS Divergence, and better action prediction (F1). These differences highlight the importance of evaluating beyond embedding similarity, considering both lexical and behavioral fidelity when assessing model imitation quality.

\begin{table*}[htbp!]
\centering
\tiny
\begin{tabular}{lccccc} 
\hline
 & \textbf{GPT 4.1-mini} & \textbf{o3-mini} & \textbf{Qwen2.5-7B-Instruct} & \textbf{Qwen2.5-7B-Instruct$_{focal\_loss}$} & \textbf{Qwen2.5-7B-Instruct$_{CE\_loss}$} \\
\hline
Jaccard Similarity (↑) & 0.0132 & 0.0207 & 0.0102 & 0.1220 & \textbf{0.1313} \\
Avg. Cosine Similarity (↑) & 0.9201 & 0.9136 & 0.9167 & \textbf{0.9581} & 0.9527\\
Max. Cosine similarity (↑) & 0.9463 & 0.9132 & 0.9126 & \textbf{1.0000} & \textbf{1.0000} \\
JS Divergence (↓) & 0.3894 & 0.4777 & 0.5109 & 
\textbf{0.2582} & 0.2611 \\
\hline
F1 Score (↑) & 0.3353 & 0.3441 & 0.3308 & 0.3425 & \textbf{0.3547} \\
\hline
\end{tabular}
\caption{\label{tab:metrics_averaged} \textbf{Population-level performance metrics} across the five clusters using \texttt{BluePrint} as ground truth.}
\end{table*}

Table~\ref{tab:metrics_averaged} summarizes performance averaged over all five clusters. Both of our fine-tuned models show substantial gains across nearly all metrics, including a 2x reduction in JS Divergence and a 10x increase in Jaccard Similarity, indicating better lexical and distributional alignment with human behavior. However, F1 scores remain comparable to the untrained baseline, suggesting that while our models improve in generating realistic content, they still struggle to reliably predict the specific actions users would take in context. This highlights an important open challenge for future work on modeling user decision-making.

\begin{figure}[htbp!]
\centering
\includegraphics[width=0.5\textwidth]{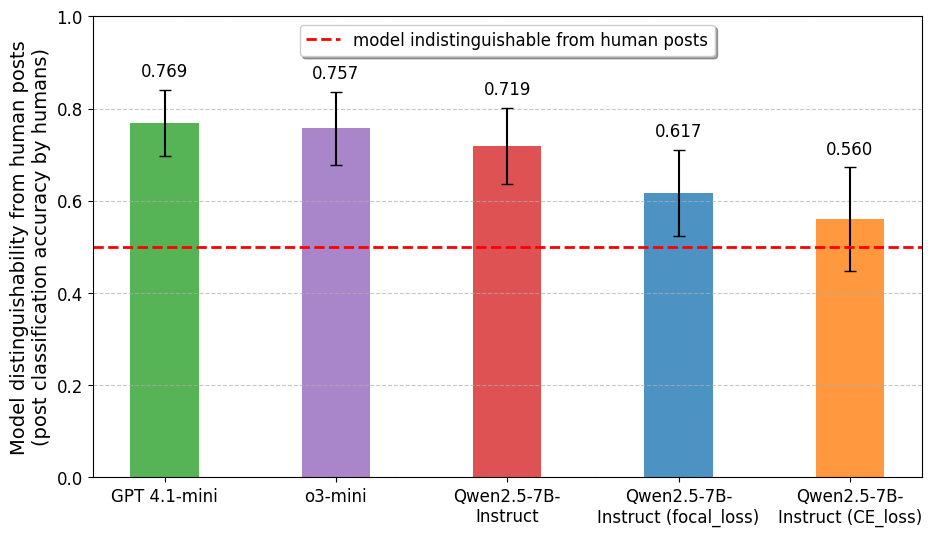}
\caption{\label{fig:human_eval_results} Human accuracy in distinguishing real from model-generated posts (lower is better; 95\% CI). A score of 0.5 represents random guessing, meaning the model is fully indistinguishable from real users. Models fine-tuned on \texttt{BluePrint} are significantly harder to identify, with our best achieving near-random distinguishability (56.0\%).}
\end{figure}

Our evaluation metrics are corroborated by human evaluation. As shown in Figure \ref{fig:human_eval_results}, models fine-tuned on \texttt{BluePrint} are significantly harder for humans to distinguish from real posts, with the best model being correctly identified only 56.0\% of the time. In contrast, the strongest untrained model is identified 71.9\% of the time. Since a perfectly human-like model should be identified at random (50\%), these results suggest that \texttt{BluePrint}-trained models substantially improve human-likeness and constitute a step towards better social media simulations, while still leaving room for future progress.
\section{Ethics Statement} \label{sec:ethics}
Training models that replicate user behaviour requires care that responsibly addresses ethical issues. We collected publicly available Bluesky data using the official Jetstream client in compliance with Bluesky's \href{https://bsky.social/about/support/tos}{Terms of Service}. While users understand their public posts are visible to others, we recognize they did not explicitly consent to research use, where ``the publicness of social media data has trivialized the question of consent'' \cite{berkeley2023privacy}. It is therefore important to preserve their privacy.

To protect user privacy, we implemented multiple, distinct and cumulative safeguards: replacing personally identifiable information (Section \ref{sec:pii_removal}), using relative rather than absolute timestamps (Section \ref{sec:relative_timestamp}), pseudonymizing usernames (Section \ref{sec:pseudonymization}), and analyzing aggregated behavioral archetypes instead of individual profiles. These measures substantially reduce re-identification risk while preserving essential data structure.

We also respect users' right to control their data by allowing them at any time to access a public \href{https://forms.gle/eydoZiHjLokXvgr88}{removal form} where Bluesky users can request deletion by submitting their DID. Upon verification, we promptly remove requested data from both our dataset and HuggingFace repository, exceeding Bluesky's standard deletion guarantees.

The dataset will be available on HuggingFace under a responsible-use license requiring agreement to terms that restrict usage to research purposes and prohibit unlawful or unethical applications. We are releasing only anonymized social media records, not derived LLMs or simulated personas.


\DatasetFramework and BluePrint enable rigorous study of LLMs as social media agents in political contexts, addressing risks of misinformation, echo chambers, and bias. Recent research demonstrates that even advanced LLMs can generate convincing election disinformation \cite{williams2025llm_disinformation}. Our dataset supports improved detection and mitigation strategies while acknowledging dual-use potential. This approach aligns with ethical frameworks for AI research emphasizing shared responsibility in governance \cite{grinbaum2024dual}.

\section{Conclusion}

\DatasetFramework and the \texttt{BluePrint} dataset establish a foundation for the systematic and ethical development of LLM-based social media agents. By structuring social media interactions as next-action prediction sequences and representing users through diverse behavioral personas, our framework enables rigorous evaluation of LLMs' capacity to simulate realistic social media behavior.

Our empirical results demonstrate that while LLMs can be fine-tuned to generate text that is linguistically consistent with target personas, they struggle to replicate broader behavioral patterns—particularly action diversity and community-specific engagement dynamics. This limitation underlines the need for more sophisticated modeling approaches that extend beyond text generation to capture both user-level and population-level interaction behaviors.

Future research directions include expanding our framework to incorporate multi-modal content, longer-term engagement patterns, and cross-platform dynamics. We anticipate that \DatasetFramework and \texttt{BluePrint} will serve as valuable resources for developing more realistic, fair, and robust social simulations, thereby advancing both AI research and computational social science.

\section*{Limitations} \label{sec:limitations}

While \DatasetFramework provides a structured and privacy-preserving approach for constructing social media user modeling datasets, several limitations remain. First, the framework captures only observable user actions, excluding content users viewed but did not engage with. It also does not distinguish between human and automated accounts, potentially introducing bot-generated noise. Additionally, generated datasets reflect the demographic and ideological biases of the source platform, which our clustering cannot fully mitigate.

\texttt{BluePrint} further focuses on a single month (March 2025) and English-language content, limiting coverage of Canada’s bilingual discourse. The dataset also likely exhibits a pronounced left-leaning bias reflective of the platform's user base during the collection period, constraining political generalizability. Finally, our heuristic linking user-directed actions to the target’s most recent post may not fully capture user intent, and results should be interpreted with this limitation in mind.


\bibliography{custom}
\appendix

\appendix
\section{Bluesky 2025 Canadian Federal Election Keywords} \label{sec:bluesky_2025_cad_federal_election_keywords}

\textbf{Handles (97)}: \scriptsize [``@aiaconomp.bsky.social'', ``@alainrayes.bsky.social'', ``@alexandramendes.bsky.social'', ``@alexboulerice.bsky.social'', ``@aliehsassi.bsky.social'', ``@alistairmacgregor.bsky.social'', ``@andreannelarouche.bsky.social'', ``@anitaanandmp.bsky.social'', ``@anitavandenbeld.bsky.social'', ``@annamgainey.bsky.social'', ``@aryacanada.bsky.social'', ``@avankoeverden.bsky.social'', ``@blakedesjarlais.bsky.social'', ``@brianmassemp.bsky.social'', ``@carolinebq.bsky.social'', ``@charlieangus104.bsky.social'', ``@chrisbittle.bsky.social'', ``@chrystia-freeland.bsky.social'', ``@coteau.bsky.social'', ``@dianelebouthillier.bsky.social'', ``@dickcannings.bsky.social'', ``@dleblancnb.bsky.social'', ``@donvdavies.bsky.social'', ``@drbrendanhanley.bsky.social'', ``@elizabethmay.bsky.social'', ``@gordjohns.bsky.social'', ``@heathermcpherson.bsky.social'', ``@hedyfry.bsky.social'', ``@honjudysgro.bsky.social'', ``@iqrakhalidmp.bsky.social'', ``@jagmeetsingh.ca``@jeanyip3.bsky.social'', ``@jennykwan.bsky.social'', ``@juliedabrusin.bsky.social'', ``@jyduclos.bsky.social'', ``@kamalkheralib.bsky.social'', ``@karinagould.bsky.social'', ``@kayabagaarielle.bsky.social'', ``@kristinamichaud.bsky.social'', ``@laurelcollins.bsky.social'', ``@leahgazan.bsky.social'', ``@leiladance.bsky.social'', ``@lindsaymathyssen.bsky.social'', ``@lisahepfner.bsky.social'', ``@lisamariebarron.bsky.social'', ``@lloydlongfield.bsky.social'', ``@loriidlout.bsky.social'', ``@lucberthold.bsky.social'', ``@maloneyj.bsky.social'', ``@marcgserremp.bsky.social'', ``@marcmillermp.bsky.social'', ``@marilenegill.bsky.social'', ``@mark-carney.bsky.social'', ``@markgerretsen.bsky.social'', ``@markhollandlib.bsky.social'', ``@martchampoux.bsky.social'', ``@maryng.bsky.social'', ``@mattjeneroux.bsky.social'', ``@mbjdepute.bsky.social'', ``@melaniejoly.bsky.social'', ``@mflalonde.bsky.social'', ``@michellegarner.bsky.social'', ``@morricemike.bsky.social'', ``@mp-bonitazarrillo.bsky.social'', ``@mpjulian.bsky.social'', ``@nathaliesinclaird.bsky.social'', ``@nikiashton.bsky.social'', ``@pamdamoff.bsky.social'', ``@pascalestonge.bsky.social'', ``@patrickbweiler.bsky.social'', ``@pattyhajdu.bsky.social'', ``@pbainsy.bsky.social'', ``@pierrepaul-hus.bsky.social'', ``@rachelbendayan.bsky.social'', ``@rachelreading.bsky.social'', ``@rboissonnault.bsky.social'', ``@rechievaldez.bsky.social'', ``@renevillemure.bsky.social'', ``@roboliphant.bsky.social'', ``@ronmckinnon.bsky.social'', ``@seanfraser.bsky.social'', ``@seblemire.bsky.social'', ``@shaunchenmp.bsky.social'', ``@sherryromanado.bsky.social'', ``@stevenguilbeault.bsky.social'', ``@stevenmackinnon``@sylvieberube.bsky.social'', ``@taleeb.bsky.social'', ``@taylorbachrach.bsky.social'', ``@terrybeech.bsky.social'', ``@tonyvanbynen.bsky.social'', ``@turnbullwhitby.bsky.social'', ``@valbradfordmp.bsky.social'', ``@viraniarif.bsky.social'', ``@vivianelapointe.bsky.social'', ``@yasirnaqvicdn.bsky.social'', ``@yfblanchet.bsky.social'']

\normalsize \textbf{Party Identifiers and Hashtags (43)}: \scriptsize [``baylis'', ``blanchet'', ``blocqc'', ``bq'', ``carney'', ``chrystia'', ``chrystiafreeland'', ``conservative'', ``cpc'', ``dhalla'', ``elizabeth'', ``elizabethmay'', ``exln45'', ``frank'', ``frankbaylis'', ``freeland'', ``gould'', ``green'', ``jagmeet'', ``jagmeetsingh'', ``jonathan'', ``jonathanpedneault'', ``justin'', ``justintrudeau'', ``karina'', ``karinagould'', ``liberal'', ``lpc'', ``mark'', ``markcarney'', ``may'', ``ndp'', ``paysqc'', ``pedneault'', ``pierre'', ``pierrepoilievre'', ``poilievre'', ``ruby'', ``rubydhalla'', ``singh'', ``trudeau'', ``yves'', ``yvesblanchet'']

\normalsize \textbf{General Political Terms (11)}: \scriptsize [``politics'', ``polcan'', ``canada'', ``canadapolitics'', ``canadian'', ``canadians'', ``canpol'', ``canpoli'', ``cdnpoli'', ``cdnpolitics'', ``election'']
\normalsize

\section{Computational Resources} \label{sec:compute_resources}
The dataset processing and clustering was performed using the Compute Canada cluster using 10 AMD EPYC 7413 (Zen 3) cpus, and one NVidia A100SXM4 (40 GB memory) GPU with 160G of RAM. Each clustering size (2, 25, 100 and 1000) took about 2 hours of compute.

We performed the privacy protection processes (PII removal and anonymization, timestamp obfuscation, and user pseudonymization) on an AWS EC2 \verb|c7i.4xlarge| instance (16 vCPUs, 32 GiB memory), and it took approximately 2 hours of compute to complete.

Models were finetuned on the Compute Canada cluster using one MD EPYC 7413 (Zen 3) cpus, and one NVidia A100SXM4 (40 GB memory) GPU with 48G of RAM. Each model took between 3 and 23 hours of compute to finetune.

The finetuned models (and the base \texttt{Qwen-2.5-7B-Instruct}) were ran on the Compute Canada cluster using one MD EPYC 7413 (Zen 3) cpus, and one NVidia A100SXM4 (40 GB memory) GPU with 16G of RAM to generate samples used to compute metrics. Generating a total of 3000 samples took cumulatively about 1 hour of compute.

\section{Qwen Fine-tuning} \label{sec:qwen_fine_tuning}
Models are tasked with next-action prediction, either by completing a missing message in a conversation thread or by generating a new standalone post. Each prompt includes a history of posts authored by users from the same behavioral cluster as the target message. We use structured JSON outputs to standardize model generations across text and action predictions.

\begin{tcolorbox}[colback=gray!10, colframe=gray!80, title=System prompt]
You are a user on social media. Your goal is to write posts and interact with other users' posts.\\

You should give your reply in a JSON format. You have the ability to interact 
with the other users' messages by either writing a message of your own or with
the following actions: like, follow, repost, ignore.\\

Here is an example of a reply:\\
\{\\
    "actions": \{\\
    "like": true,\\
    "follow": false,\\
    "repost": false,\\
    "ignore": false\\
  \},\\
  "text": "This is a sample reply to the user's message."\\
\}\\

Writing the 'text' field is optional if you are replying to an other user's post.
If you are writing your own post, you should not include the 'actions' field.\\

--- FEED HISTORY ---\\
{{history}}\\
--- END OF FEED HISTORY ---
\end{tcolorbox}

\begin{figure*}
\begin{tcolorbox}[colback=cyan!10, colframe=cyan!80, title=Post schema,width=\textwidth]
\begin{lstlisting}
{
    "$schema": "http://json-schema.org/draft-04/schema",
    "description": "",
    "type": "object",
    "properties": {
        "text": {
            "type": "string"
        }
    },
    "required": [
        "text"
    ],
    "additionalProperties": false
}
\end{lstlisting}
\end{tcolorbox}
\end{figure*}

\begin{figure*}
\begin{tcolorbox}[colback=cyan!10, colframe=cyan!80, title=Reply schema, width=\textwidth]
\begin{lstlisting}
{
    "$schema": "http://json-schema.org/draft-04/schema",
    "description": "",
    "type": "object",
    "properties": {
        "actions": {
            "type": "object",
            "properties": {
                "like": {
                    "type": "boolean"
                },
                "follow": {
                    "type": "boolean"
                },
                "repost": {
                    "type": "boolean"
                },
                "ignore": {
                    "type": "boolean"
                }
            },
            "required": [
                "like",
                "follow",
                "repost",
                "ignore"
            ],
            "additionalProperties": false
        },
        "text": {
            "type": "string"
        }
    },
    "required": [
        "actions",
        "text"
    ],
    "additionalProperties": false
}
\end{lstlisting}
\end{tcolorbox}
\end{figure*}

Given the limited capability of 7B-parameter models compared to proprietary models, we observed occasional failures such as gibberish outputs or broken JSON formatting. To mitigate this, we generate three candidate responses per prompt and ask \texttt{Qwen-2.5-7B-Instruct} to select the most human-like response. Post-processing steps include filtering out invalid characters and truncating noisy outputs.

To preserve the instruction-following capabilities of \texttt{Qwen-2.5-7B-Instruct} during fine-tuning, we augment \texttt{BluePrint} data with general-purpose instruction-following examples from \texttt{databricks/databricks-dolly-15k} \cite{DatabricksBlog2023DollyV2} and context-specific examples generated following best practices in instruction tuning \cite{cheng2024adaptinglargelanguagemodels}. While \texttt{Qwen3-7B} was released during our experiments, preliminary tests showed no performance gains over \texttt{Qwen-2.5-7B-Instruct}, and we therefore proceeded with the more stable 2.5 version.

\section{Human Eval Experimental setting} \label{sec:human_eval}

\begin{figure*}
\begin{tcolorbox}[colback=gray!10, colframe=gray!80, title=Social Media AI Turing Test: Instructions presented to the participants, width=\textwidth]
Welcome to our experiment! We're testing how well AI models can imitate real social media users.

You'll be shown pairs of tweets — one written by a real person and one generated by AI.

\textbf{Your task:} Guess which tweet was AI-generated and which one was written by a human.

\textbf{How it works:}
\begin{itemize}
  \item You'll see a conversation thread (if any) followed by two tweets labeled A and B
  \item One tweet is from a real person, one is AI-generated
  \item Click on the tweet you think was AI-generated
  \item If you can't decide, you can choose ``I'm not sure''
  \item Continue as long as you'd like — every response helps our research!
\end{itemize}
\end{tcolorbox}
\end{figure*}

\begin{figure*}[hbtp!]
    \centering
    \includegraphics[width=\linewidth]{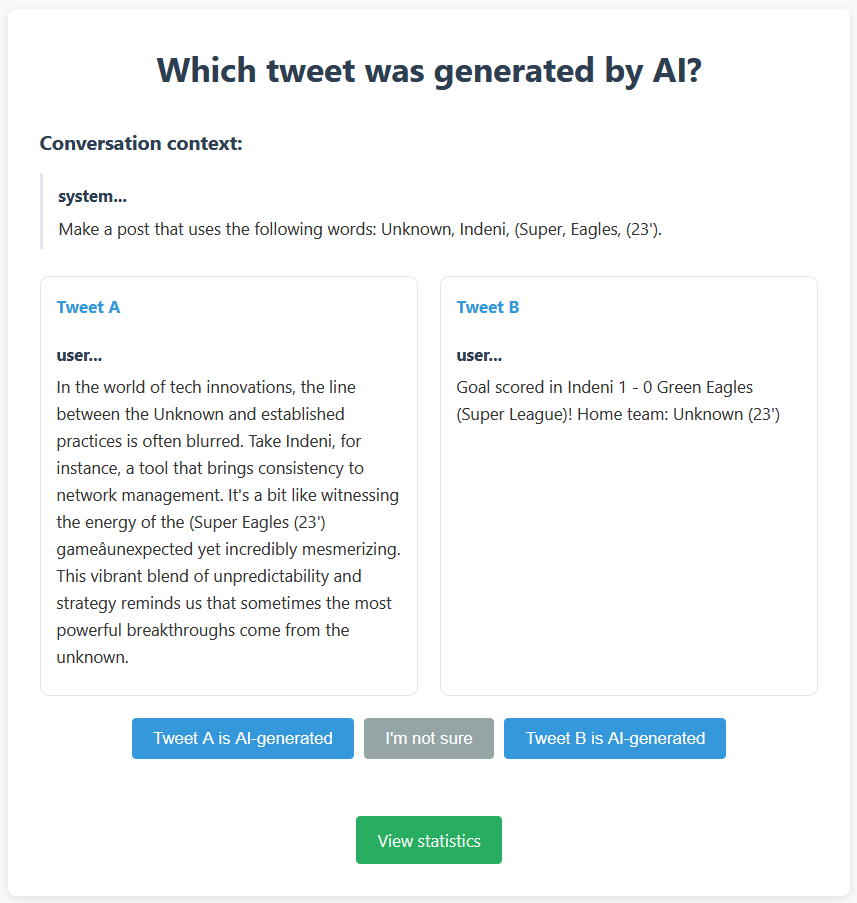}
    \caption{\label{fig:appendix_human_eval_layout} Layout of the questions as presented to participants}
\end{figure*}

\section{Cluster statistics} \label{sec:cluster_stats}

\begin{table*}[hbtp!]
\centering
\begin{tabular}{lrrlrr}
\hline
Action & Cluster 0 & Cluster 1 & Average ± std & Max & Total \\
\hline
posts & 500147 & 973204 & 736675.50 ± 236528.50 & 973204 & 1473351 \\
replies & 19506 & 54119 & 36812.50 ± 17306.50 & 54119 & 73625 \\
quote & 15345 & 27067 & 21206.00 ± 5861.00 & 27067 & 42412 \\
post\_update & 914 & 86 & 500.00 ± 414.00 & 914 & 1000 \\
post\_delete & 8 & 15 & 11.50 ± 3.50 & 15 & 23 \\
repost & 285 & 993 & 639.00 ± 354.00 & 993 & 1278 \\
unrepost & 3 & 1 & 2.00 ± 1.00 & 3 & 4 \\
like & 269942 & 1318815 & 794378.50 ± 524436.50 & 1318815 & 1588757 \\
unlike & 2011 & 8914 & 5462.50 ± 3451.50 & 8914 & 10925 \\
follow & 988316 & 2529533 & 1758924.50 ± 770608.50 & 2529533 & 3517849 \\
unfollow & 27916 & 41153 & 34534.50 ± 6618.50 & 41153 & 69069 \\
block & 115 & 690 & 402.50 ± 287.50 & 690 & 805 \\
unblock & 0 & 2 & 1.00 ± 1.00 & 2 & 2 \\
\hline
total & 1824508 & 4954592 & 3389550.00 ± 1565042.00 & 4954592 & 6779100 \\
\hline
users & 76168 & 160163 & 118165.50 ± 41997.50 & 160163 & 236331 \\
\hline
\end{tabular}
\caption{Statistics for size 2 clustering}
\end{table*}

\begin{table*}[hbtp!]
\centering
\small
\begin{tabular}{lrrrrrlrr}
\hline
Action & Cluster 0 & Cluster 1 & ... & Cluster 23 & Cluster 24 & Average ± std & Max & Total \\
\hline
posts & 34101 & 15686 & ... & 72011 & 79036 & 61960.16 ± 40726.10 & 155252 & 1549004 \\
replies & 1794 & 220 & ... & 1609 & 2710 & 1876.52 ± 2660.90 & 12908 & 46913 \\
quote & 224 & 123 & ... & 1001 & 3480 & 1696.48 ± 2314.31 & 10075 & 42412 \\
post\_update & 50 & 2 & ... & 145 & 57 & 40.00 ± 76.47 & 357 & 1000 \\
post\_delete & 0 & 0 & ... & 1 & 3 & 0.92 ± 0.98 & 3 & 23 \\
repost & 13 & 0 & ... & 44 & 74 & 51.12 ± 47.77 & 155 & 1278 \\
unrepost & 0 & 0 & ... & 0 & 0 & 0.16 ± 0.37 & 1 & 4 \\
like & 10155 & 2579 & ... & 28130 & 108601 & 63550.28 ± 58909.05 & 195547 & 1588757 \\
unlike & 79 & 20 & ... & 245 & 780 & 437.00 ± 400.97 & 1268 & 10925 \\
follow & 51126 & 4223 & ... & 185512 & 166658 & 140713.96 ± 124800.60 & 531633 & 3517849 \\
unfollow & 1712 & 226 & ... & 6772 & 2688 & 2762.76 ± 1922.79 & 7510 & 69069 \\
block & 3 & 6 & ... & 8 & 42 & 32.20 ± 30.65 & 134 & 805 \\
unblock & 0 & 0 & ... & 0 & 1 & 0.08 ± 0.27 & 1 & 2 \\
\hline
total & 99257 & 23085 & ... & 295478 & 364130 & 273121.64 ± 212985.39 & 838465 & 6828041 \\
\hline
users & 6700 & 3062 & ... & 8901 & 11517 & 9453.24 ± 4194.91 & 18452 & 236331 \\
\hline
\end{tabular}
\caption{Statistics for size 25 clustering}
\end{table*}

\begin{table*}[hbtp!]
\centering
\small
\begin{tabular}{lrrrrrlrr}
\hline
Action & Cluster 0 & Cluster 1 & ... & Cluster 98 & Cluster 99 & Average ± std & Max & Total \\
\hline
posts & 62823 & 14852 & ... & 2689 & 86091 & 15725.92 ± 16779.41 & 86091 & 1572592 \\
replies & 2890 & 333 & ... & 87 & 8317 & 390.15 ± 912.60 & 8317 & 39015 \\
quote & 4707 & 475 & ... & 22 & 6470 & 424.12 ± 975.55 & 6470 & 42412 \\
post\_update & 157 & 3 & ... & 0 & 17 & 10.00 ± 28.49 & 161 & 1000 \\
post\_delete & 0 & 0 & ... & 0 & 1 & 0.23 ± 0.53 & 3 & 23 \\
repost & 52 & 15 & ... & 0 & 94 & 12.78 ± 18.39 & 98 & 1278 \\
unrepost & 0 & 0 & ... & 0 & 0 & 0.04 ± 0.20 & 1 & 4 \\
like & 54032 & 23084 & ... & 901 & 132631 & 15887.57 ± 22213.23 & 132631 & 1588757 \\
unlike & 420 & 134 & ... & 11 & 882 & 109.25 ± 150.82 & 882 & 10925 \\
follow & 101995 & 30562 & ... & 3875 & 61182 & 35178.49 ± 46442.02 & 305560 & 3517849 \\
unfollow & 2881 & 480 & ... & 171 & 631 & 690.69 ± 813.06 & 4206 & 69069 \\
block & 12 & 11 & ... & 0 & 24 & 8.05 ± 10.45 & 49 & 805 \\
unblock & 0 & 1 & ... & 0 & 0 & 0.02 ± 0.14 & 1 & 2 \\
\hline
total & 229969 & 69950 & ... & 7756 & 296340 & 68437.31 ± 81704.65 & 466196 & 6843731 \\
\hline
users & 4869 & 3192 & ... & 711 & 4911 & 2363.31 ± 1575.47 & 7518 & 236331 \\
\hline
\end{tabular}
\caption{Statistics for size 100 clustering}
\end{table*}

\begin{table*}[hbtp!]
\centering
\small
\begin{tabular}{lrrrrrlrr}
\hline
Action & Cluster 0 & Cluster 1  & ... & Cluster 998 & Cluster 999 & Average ± std & Max & Total \\
\hline
posts & 61 & 16013 & ... & 1483 & 78 & 1592.61 ± 3396.10 & 40808 & 1592612 \\
replies & 0 & 433 & ... & 42 & 2 & 31.83 ± 110.78 & 1900 & 31829 \\
quote & 0 & 481 & ... & 91 & 0 & 42.41 ± 182.12 & 2817 & 42412 \\
post\_update & 0 & 0 & ... & 0 & 0 & 1.00 ± 6.57 & 127 & 1000 \\
post\_delete & 0 & 0 & ... & 0 & 0 & 0.02 ± 0.16 & 2 & 23 \\
repost & 0 & 7 & ... & 0 & 0 & 1.28 ± 3.67 & 56 & 1278 \\
unrepost & 0 & 0 & ... & 0 & 0 & 0.00 ± 0.06 & 1 & 4 \\
like & 66 & 16043 & ... & 854 & 33 & 1588.76 ± 4051.27 & 60181 & 1588757 \\
unlike & 0 & 168 & ... & 4 & 0 & 10.93 ± 27.13 & 389 & 10925 \\
follow & 124 & 10856 & ... & 2346 & 10 & 3517.85 ± 7773.67 & 74867 & 3517849 \\
unfollow & 0 & 99 & ... & 25 & 0 & 69.07 ± 178.74 & 2064 & 69069 \\
block & 0 & 11 & ... & 0 & 0 & 0.81 ± 2.21 & 26 & 805 \\
unblock & 0 & 0 & ... & 0 & 0 & 0.00 ± 0.04 & 1 & 2 \\
\hline
total & 251 & 44111 & ... & 4845 & 123 & 6856.56 ± 14430.48 & 146908 & 6856565 \\
\hline
users & 25 & 1307 & ... & 261 & 26 & 236.33 ± 279.22 & 1715 & 236331 \\
\hline
\end{tabular}
\caption{Statistics for size 1000 clustering}
\end{table*}

\section{TF-IDF Keywords for size 25 clustering}

\begin{table*}[htbp!]
\centering
\small
\begin{tabular}{lll}
\hline
\textbf{Cluster} & \textbf{Label} & \textbf{Top Words}\\
\hline
0 & Scientists & 2503, arxiv, abs, 1012, diagnosed \\
1 & Sports & nba, ras, aewrevolution, qb, nct \\
2 & Politics: U.S. intelligence & halper, pacification, rubio, krasnov, scotus \\
3 & Pop culture: Eastern Media & vtuber, pokemon, initiator, sonnets \\
4 & Millenials & 1012, diagnosed, amputated, stingy, gofund \\
5 & Religion & padme, godswill, mani, hum, wixsite \\
6 & Politics: SCOTUS criticism & dictator, scotus, llp, corruption, amendment \\
7 & Politics: Concern for US Democracy & ssnews, dictator, demise, rigging, midterms \\
8 & Politics: Florida & floridaelection, gayvalimont, avanti, cheeto, activism \\
9 & Politics: Anit-conservative Canada & canadasky, canadastrong, neverpoilievre, abpoli, uspoli \\
10 & Politics: Liberal Party & lpc, danielle, neverpoilievre, freeland, canadastrong \\
11 & Politics: U.S. Congressional Decorum & moskowitz, houlahan, himes, decorum, costa \\
12 & Politics: Economics & tradewar, uspoli, neverpoilievre, politic, imports \\
13 & Pop culture: Western Media & ruffalo, superman, innie, thearchers, romance \\
14 & Politics: Canadian Politics & vive, danielle, dictator, charlieangus104, freeland \\
15 & Pop culture: Severance (TV Series) & innie, iqjl, 24h, outie, lumon \\
16 & Politics: Conspiracy Theories & jews, pip, vaccines, legitimate, nhs \\
17 & Politics: Canada-U.S import/exports & annexation, dairy, potash, export, dictator \\
18 & Pop culture: AO3 (Archive of Our Own) & fic, ocs, fanart, lore, 7250 \\
19 & Politics: Left-leaning Politics & leftists, progressives, socialist, liberalism, libs \\
20 & GenZ: NSFW & arabic\_word, hehe, arabic\_word\_2, ashamed, lick \\
21 & GenZ: Green (the color) &  thewildimages, yuri, prefecture, pinched, grapes \\
22 & Creative Work & deviantart, precip, otd, serene, iembot \\
23 & Politics: Florida & gayforcongre, valimont, mobilize, crawfordforcourt, stopshadyschimel \\
24 & Politics: Canadian inter-provincial politics & railway, provinces, provincial, annexation, never51 \\
\hline
\end{tabular}
\caption{\label{tab:appendix_tf_idf_keywords} Top 5 TF-IDF keywords by cluster (size 25 clustering)}
\end{table*}

\textbf{Interpretation Note.} The listed topics reflect the most salient themes discussed by users in each cluster based on TF-IDF keyword analysis. However, as users typically engage with a wide range of topics, not all messages within a cluster are strictly related to the provided label. The labels should be understood as broad characterizations of the cluster’s average topical focus, rather than exhaustive or exclusive descriptors of all cluster content.

Additionally, while TF-IDF is a useful tool for surfacing characteristic terms, it is not without limitations. In particular, the method can be sensitive to anomalous posting behavior, such as a single high-volume account influencing the top terms. While such cases are rare, they may account for occasional mismatches between the assigned label and the listed keywords.

To support further exploration and validation, full keyword lists, cluster-level statistics, and representative sample posts (medoids) are provided in the publicly available dataset release on Hugging Face.

\begin{table*}[hbtp!]
\centering
\tiny
\begin{tabular}{llp{1.5cm}p{1.5cm}p{2cm}p{2cm}p{2cm}}
\hline
\textbf{Cluster} & \textbf{Metric} & \textbf{GPT 4.1-mini} & \textbf{o3-mini} & \textbf{Qwen2.5-7B-Instruct} & \textbf{Qwen2.5-7B-Instruct$_{focal\_loss}$} & \textbf{Qwen2.5-7B-Instruct$_{CE\_loss}$} \\
\hline

\multirow{5}{*}{Cluster 6}
  & Jaccard Similarity      & 0.0050 & 0.0152 & 0.0050 & 0.0989 & \textbf{0.1429} \\
  & Avg. Cosine Similarity  & 0.8944 & 0.8896 & 0.8907 & \textbf{0.9293} & 0.9152 \\
  & Max. Cosine Similarity   & \textbf{1.0000} & \textbf{1.0000} & \textbf{1.0000} & \textbf{1.0000} & \textbf{1.0000} \\
  & JS Divergence & 0.5340 & 0.5818 & 0.5977 & \textbf{0.4770} & 0.4973 \\
  & F1 Score                & \textbf{0.2077} & \textbf{0.2077} & 0.1913 & 0.2022 & 0.1257 \\
\hline

\multirow{5}{*}{Cluster 21}
  & Jaccard Similarity      & 0.0101 & 0.0101 & 0.0050 & 0.0753 & \textbf{0.0811} \\
  & Avg. Cosine Similarity  & 0.9137 & 0.9057 & 0.9123 & 0.9420 & \textbf{0.9465} \\
  & Max. Cosine Similarity   & \textbf{1.0000} & 0.8543 & \textbf{1.0000} & \textbf{1.0000} & \textbf{1.0000} \\
  & JS Divergence & 0.3465 & 0.4243 & 0.4617 & 0.2905 & \textbf{0.2591} \\
  & F1 Score                & 0.3000 & \textbf{0.3111} & 0.3000 & 0.2722 & 0.3056 \\
\hline

\multirow{5}{*}{Cluster 1}
  & Jaccard Similarity      & 0.0101 & 0.0000 & 0.0050 & \textbf{0.1364} & 0.1299 \\
  & Avg. Cosine Similarity  & 0.9390 & 0.9321 & 0.9341 & 0.9909 & \textbf{0.9947} \\
  & Max. Cosine Similarity   & 0.8685 & 0.8660 & 0.8581 & \textbf{1.0000} & \textbf{1.0000} \\
  & JS Divergence & 0.3322 & 0.4739 & 0.5482 & 0.0680 & \textbf{0.0435} \\
  & F1 Score                & 0.5276 & 0.5354 & 0.5276 & 0.5354 & \textbf{0.6220} \\
\hline

\multirow{5}{*}{Cluster 0}
  & Jaccard Similarity      & 0.0256 & 0.0256 & 0.0101 & \textbf{0.1299} & 0.1050 \\
  & Avg. Cosine Similarity  & 0.9225 & 0.9150 & 0.9166 & \textbf{0.9623} & 0.9485 \\
  & Max. Cosine Similarity   & \textbf{1.0000} & 0.8455 & 0.8443 & \textbf{1.0000} & \textbf{1.0000} \\
  & JS Divergence & 0.4092 & 0.4926 & 0.5230 & \textbf{0.2311} & 0.2789 \\
  & F1 Score                & 0.2866 & 0.2988 & 0.2866 & 0.3415 & \textbf{0.3720} \\
\hline

\multirow{5}{*}{Cluster 18}
  & Jaccard Similarity      & 0.0152 & 0.0526 & 0.0256 & 0.1696 & \textbf{0.1976} \\
  & Avg. Cosine Similarity  & 0.9310 & 0.9257 & 0.9297 & \textbf{0.9662} & 0.9586 \\
  & Max. Cosine Similarity   & 0.8629 & \textbf{1.0000} & 0.8605 & \textbf{1.0000} & \textbf{1.0000} \\
  & JS Divergence & 0.3252 & 0.4161 & 0.4240 & \textbf{0.2244} & 0.2267 \\
  & F1 Score                & 0.3548 & \textbf{0.3677} & 0.3484 & 0.3613 & 0.3484 \\
\hline
\end{tabular}
\caption{\label{tab:metrics_for_all_cluster_table} \textbf{Cluster-level performance metrics} for multiple models imitating social media users, using \texttt{BluePrint} as ground truth. Results are shown for five representative clusters (0, 1, 6, 18, 21) from the 25-cluster partition. }
\end{table*}

\section{Visualization of Personas}

\begin{figure*}[htbp!]
\centering
\includegraphics[width=\textwidth]{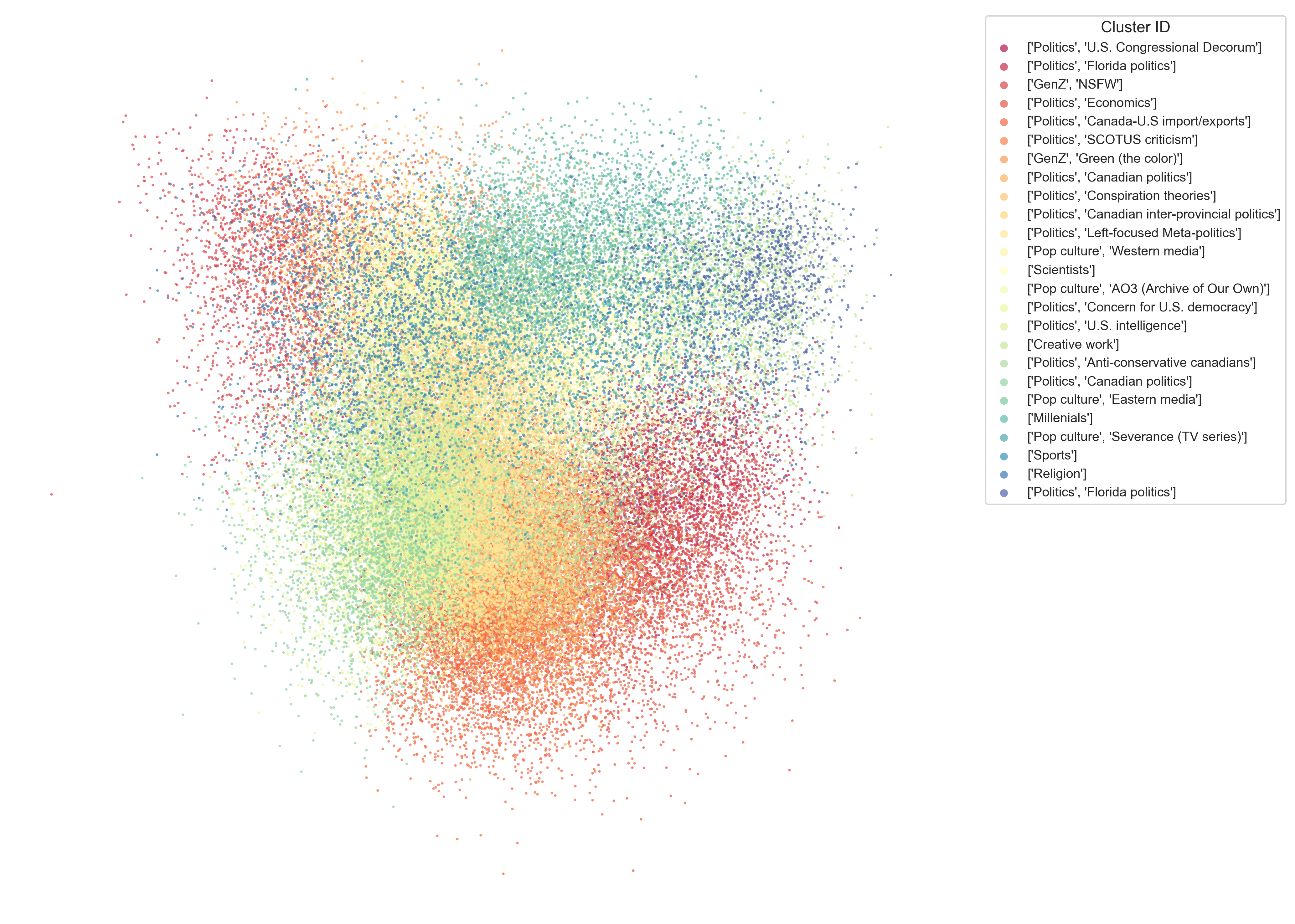}
\caption{\label{fig:cluster_viz} Visualization of user embeddings (average of all of a user's posts' embedding vectors) of 80,000 randomly selected users.}
\end{figure*}

\end{document}